\documentclass[sigconf]{acmart}
\usepackage{graphicx}
\usepackage{subcaption}
\usepackage{multirow}
\AtBeginDocument{%
  }
\setcopyright{none}
\renewcommand\footnotetextcopyrightpermission[1]{}
\begin{document}


\title{Multi-Level Modeling of Large Language Model Inference Latency and Energy via Hybrid Analytical--Machine-Learning Predictors
}


\author{Saeid Shokoufa}
\affiliation{%
  \institution{University of Southern California}
  \city{Los Angeles}
  \state{California}
  \country{USA}
}
\email{shokoufa@usc.edu}

\author{Mohammad Erfan Sadeghi}
\affiliation{%
  \institution{University of Southern California}
  \city{Los Angeles}
  \state{California}
  \country{USA}
}
\email{sadeghim@usc.edu}

\author{Mehdi Kamal}
\affiliation{%
  \institution{University of Southern California}
  \city{Los Angeles}
  \state{California}
  \country{USA}
}
\email{mehdi.kamal@usc.edu}

\author{Massoud Pedram}
\affiliation{%
  \institution{University of Southern California}
  \city{Los Angeles}
  \state{California}
  \country{USA}
}
\email{pedram@usc.edu}

\renewcommand{\shortauthors}{Trovato et al.}

\begin{abstract}
The rapid scaling of Large Language Models (LLMs) has significantly increased computational cost, energy consumption, and inference latency, making accurate estimation essential for sustainable artificial intelligence deployment and hardware-aware design. In this work, we introduce Hybrid Modeling for Energy and Latency of LLMs (HYMELL), a hybrid three-level framework for estimating LLM inference latency and energy by combining analytical modeling with machine learning (ML).
HYMELL models LLM execution through a three-level hierarchy: analytical estimation of primitive operations, ML prediction of higher-level components, and an end-to-end model that captures system-level overheads across both prefill and decode phases. The framework supports diverse architectures, including dense and mixture-of-experts (MoE) feed-forward networks (FFNs), as well as multi-head attention (MHA) and grouped-query attention (GQA) mechanisms.
Evaluated on an NVIDIA H100 graphics processing unit (GPU), HYMELL achieves high predictive accuracy; notably, for LLaMA 3 8B, it attains less than 5\% error for both prefill and decode phases. By predicting execution costs directly from architectural parameters, it enables fast, hardware-free design space exploration and energy-efficient optimization.

\end{abstract}





\maketitle
\pagestyle{plain}
\section{INTRODUCTION}

Neural networks (NNs), particularly large language models (LLMs), 
have become central to modern AI. Most LLMs are based on the 
Transformer architecture introduced in \textit{Attention Is All You Need} 
\cite{vaswani2017attention} and are commonly implemented as either dense 
or mixture-of-experts (MoE) models \cite{fedus2022switch}. Their rapid 
scaling and deployment, however, have introduced substantial energy costs 
\cite{schwartz2019greenai, luccioni2024power}; for example, training GPT-3 
was estimated to consume about 1,287 MWh of electricity and produce 
552 tCO$_2$e \cite{patterson2021carbon}. 
Consequently, extensive research has focused on reducing NN and LLM energy 
through techniques such as pruning, quantization, efficient attention, 
sparsity, and memory-efficient serving 
\cite{han2016deepcompression, abbasi2026integration,strumolo2025data, frantar2022gptq, dettmers2022llmint8, dao2022flashattention, dao2023flashattention2fasterattentionbetter, frantar2023sparsegpt, kwon2023efficient}.
NNs, and especially LLMs, are now used across a wide range of applications, 
including natural-language processing, engineering, healthcare, 
finance, law, and education 
\cite{chen2021evaluating,abdollahi2026unified, abdollahi2026hdlforgetwostagemultiagentframework,golkarieh2025semi,fayyazi2026coft,khezresmaeilzadeh2025preserving, singhal2023large, wu2023bloomberggpt, guha2023legalbench, kasneci2023chatgpt}.
Their rapidly growing adoption makes efficient 
and accurate modeling of LLM inference latency and energy increasingly 
important.

\begin{figure}[t]
\centering
\includegraphics[scale=0.43]{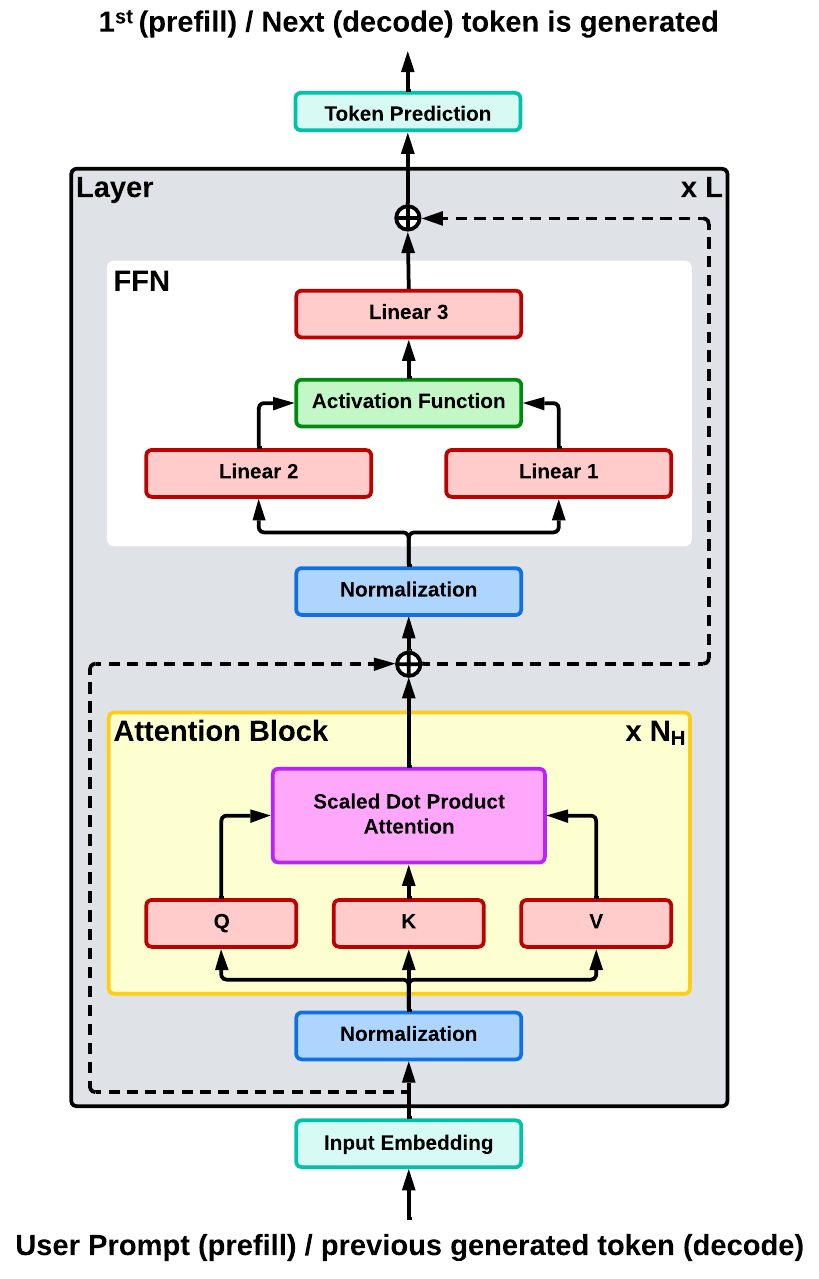}
\caption{General architecture of an LLM ($L$ and $N_H$ represent the number of layers and attention heads, respectively).}
\label{fig:llm_arch}
\end{figure}

As shown in Figure~\ref{fig:llm_arch}, despite implementation differences, most LLMs share a common architecture of stacked layers comprising attention mechanisms (multi-head attention (MHA) or grouped-query attention (GQA)) and feed-forward networks (FFNs; dense or mixture-of-experts (MoE)). At a lower level of execution, these components map to general matrix multiplications (GEMMs) and element-wise or nonlinear operations (e.g., normalization, softmax). While GEMMs dominate prefill computation, memory-bound element-wise operations account for a significant share of total energy, especially during autoregressive decoding.



LLM inference consists of two phases with distinct characteristics. In the \textbf{prefill} phase, the full input sequence is processed in parallel, making it largely compute-bound due to large matrix multiplications. In contrast, the \textbf{decode} phase generates tokens autoregressively, with increasing reliance on key-value (KV) cache accesses, making it latency-sensitive and often memory-bound. These differences motivate separate modeling of performance and energy across phases.

In this paper, we introduce Hybrid Modeling for Energy and Latency of LLMs (HYMELL), a hybrid model for energy and latency during LLM inference. HYMELL adopts a three-stage modeling approach:
\begin{enumerate}
    \item Operator-level modeling: Analytical models first estimate the latency and energy costs of primitive operations, including normalization, softmax, and GEMM.
    \item Block-level modeling: These estimates are then used by machine learning (ML) models to predict the cost of block-level components, including attention and feed-forward network (FFN) blocks.    
    \item System-level modeling: Finally, an ML model aggregates the predicted block costs together to estimate the end-to-end inference latency and energy consumption of the full LLM.
\end{enumerate}
Our goal is to build accurate and interpretable predictors that generalize across sequence lengths and batching regimes, enabling energy-aware deployment and design decisions for both dense and sparse MoE-based LLMs.

\section{RELATED WORK}
A growing body of work has explored methods for predicting the runtime (latency) and energy consumption of LLMs.

AMALI (Analytical Model for Accurately Modeling LLM Inference) \cite{cao2025amali} proposes a detailed analytical model for LLM inference on modern graphics processing units (GPUs) by explicitly capturing low-level architectural behavior. While it achieves high accuracy, it relies on extensive manual modeling and calibration, requiring detailed per-kernel and per-workload analysis. This strong dependence on GPU microarchitectural knowledge makes the approach complex and less accessible. Moreover, adapting the model to new LLM architectures or kernels requires repeated analysis, limiting scalability.

LIFE \cite{patwari2025forecastingllminferenceperformance} introduces a hardware-agnostic analytical framework that models operator-level compute and memory behavior. However, it depends on explicit hardware specifications, such as compute throughput and memory bandwidth, as well as efficiency assumptions. Adapting the framework to new models or optimizations requires manual reconfiguration of its components. In addition, its reliance on a simulation pipeline and a configuration-driven workflow increases system complexity and limits deployment ease.

SweetSpot \cite{cavagna2026sweetspotanalyticalmodelpredicting} presents an analytical model to estimate the energy efficiency of LLM inference based on the complexity of computation and memory access. Although effective, it requires deriving detailed equations tied to transformer internals and workload structure. This limits flexibility when adapting to new models or unseen behaviors, as extending the approach requires manual reformulation. Furthermore, its structured analytical pipeline adds modeling overhead compared to more data-driven approaches.

The work in \cite{krupp2026takinglonginvestigating} investigates the use of inference time as a proxy to estimate energy consumption in API-based LLMs. However, this relies on the assumption that latency directly correlates with energy, which may not hold under varying hardware utilization, parallelism strategies, or system-level overheads. As a result, the method cannot capture fine-grained architectural or workload-dependent effects, providing only coarse-grained estimates.

\vspace{-3mm}

\section{METHODOLOGY}
\begin{figure*}[t]
\centering
\includegraphics[width=0.86\textwidth]{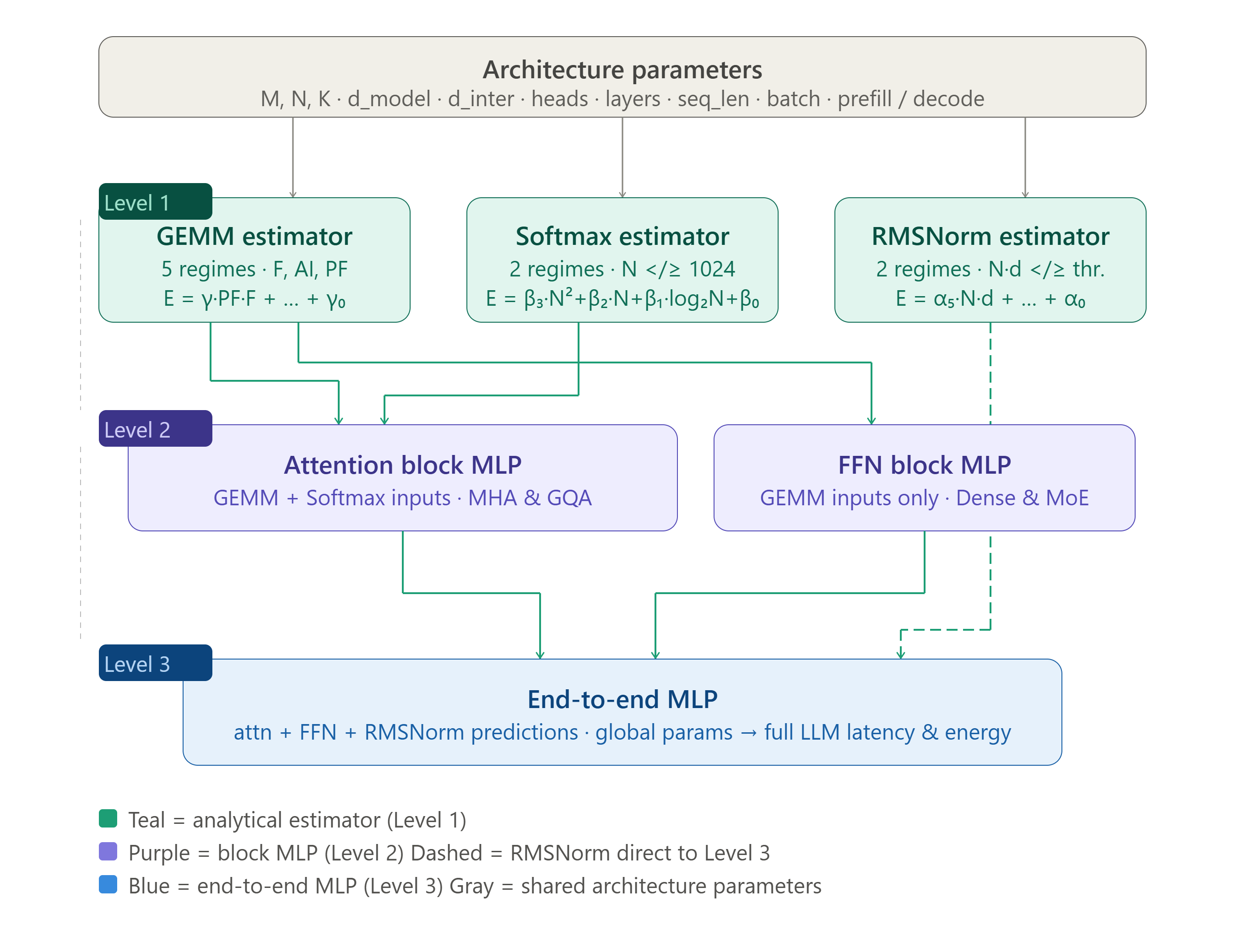}
\caption{Overview of the HYMELL hierarchical prediction framework. Level 1 analytically models primitive GPU operators. Level 2 predicts block-level execution using lightweight MLPs. Level 3 combines block-level estimates with architectural parameters to predict end-to-end latency and energy.}
\Description{A three-level HYMELL hierarchy connecting primitive GPU operator estimators, block-level MLP predictors, and an end-to-end latency and energy predictor.}
\label{fig:hymell_tree_final}
\end{figure*}

Figure~\ref{fig:hymell_tree_final} summarizes the HYMELL framework. HYMELL uses a three-level hierarchy to predict LLM inference latency and energy. Level 1 models primitive operators, including GEMM, Softmax, and RMSNorm, using analytical estimators based on operator dimensions and regime-specific features. Level 2 feeds these primitive predictions, together with architectural parameters, into lightweight MLPs to estimate attention and FFN block costs, capturing residual overheads such as reshaping, masking, activation, and routing. Level 3 combines the block-level predictions with global model parameters, including layers, sequence length, batch size, and inference mode, to estimate end-to-end latency and energy for both prefill and decode.

\subsection{RMSNorm Estimator (Analytical)}
\label{subsec:normalization}
Normalization is a lightweight operation applied independently to each token in every LLM layer. Although each invocation is inexpensive, repeated use across layers and tokens incurs non-negligible latency and energy consumption, especially during autoregressive decoding.

In modern transformer-based LLMs, RMSNorm has largely replaced traditional layer normalization (LayerNorm). Therefore, in this work, we model normalization using RMSNorm. Nevertheless, the proposed estimation methodology can be easily extended to other normalization variants, as they exhibit similar reduction- and element-wise execution patterns.

To capture execution behavior, we adopt a two-regime estimator that distinguishes between launch-bound and memory-bound execution based on the total number of processed elements. For each regime, execution time and energy are modeled as:
\begin{equation}
\begin{aligned}
E_{\text{RMSNorm}} = \alpha_5^{E}\, N\cdot d_{\text{model}}  
    + \alpha_4^{E}\, N\cdot\log_2(d_{\text{model}}) \\
    + \alpha_3^{E}\, d_{\text{model}}
    + \alpha_2^{E}\, N
    + \alpha_1^{E}\, \log_2(d_{\text{model}})
    + \alpha_0^{E}
\end{aligned}
\end{equation}
\begin{equation}
\begin{aligned}
T_{\text{RMSNorm}} = \alpha_5^{T}\, N\cdot d_{\text{model}}  
    + \alpha_4^{T}\, N\cdot\log_2(d_{\text{model}}) \\
    + \alpha_3^{T}\, d_{\text{model}}
    + \alpha_2^{T}\, N
    + \alpha_1^{T}\, \log_2(d_{\text{model}})
    + \alpha_0^{T}
\end{aligned}
\end{equation}

\noindent where $N$ is the number of tokens and $d_{\text{model}}$ is the hidden dimension, and $\alpha_i^{E}$ and $\alpha_i^{T}$ are regression coefficients. The execution regime is determined by the total workload: when $N \cdot d_{\text{model}}$ is below a threshold $th_{\text{RMS}}$, the operation is launch-bound; otherwise, it is memory-bound. This separation improves modeling accuracy, particularly for small problem sizes. 
Each term in the estimator captures a specific physical aspect: $d_{\text{model}}$ represents per-token operations and vector-length-dependent memory accesses; $N$ reflects the linear scaling with the sequence length; $N \cdot d_{\text{model}}$ serves as a proxy for total compute and dominant memory traffic; $\log_2(d_{\text{model}})$ and $N \cdot \log_2(d_{\text{model}})$ model the depth and aggregate cost of tree-based reductions; and the bias accounts for fixed kernel launch and synchronization overheads.

This two-regime formulation captures distinct execution behaviors without increasing model complexity, reducing error for small tensors while maintaining accuracy at scale.

Overall, normalization is best characterized as a memory-bound operator with non-negligible fixed overhead. Consequently, explicitly modeling normalization as a base operator yields more accurate time and energy estimates than approximating its cost using aggregate FLOP-based metrics.

\subsection{Softmax Base Estimator (Analytical)}
\label{subsec:softmax}

Softmax is a core operator in the self-attention mechanism that transforms attention scores into normalized attention weights. In LLMs, it is applied to an $N \times N$ attention matrix under a causal mask, ensuring each token attends only to previous tokens. Despite its simplicity, softmax is invoked across layers and attention heads, leading to substantial memory traffic and non-negligible latency and energy cost.

The Softmax execution exhibits two distinct behaviors depending on the length of the sequence $N$. Similar to RMSNorm, we adopt a two-regime estimator to capture launch-bound and memory-bound execution. For each regime, execution time and energy are modeled as:
\begin{equation}
\begin{aligned}
E_{\text{Softmax}} = \beta_3^{E}\, N^2
+ \beta_2^{E}\, N
+ \beta_1^{E}\, \log_2(N)
+ \beta_0^{E}\
\end{aligned}
\end{equation}
\begin{equation}
\begin{aligned}
T_{\text{Softmax}} = \beta_3^{T}\, N^2
+ \beta_2^{T}\, N
+ \beta_1^{T}\, \log_2(N)
+ \beta_0^{T}\
\end{aligned}
\end{equation}

\noindent where $N$ is the number of input tokens, and $\beta_i^{E}$ and $\beta_i^{T}$ denote the regression coefficients for energy and latency, respectively. The execution regime is determined by $N$: if $N < th_{\text{Softmax}}$, the operation is launch-bound; otherwise, it is memory-bound.
 
The kernel-launch-dominated regime corresponds to small $N$, where kernel-launch latency, masking logic, and synchronization overhead dominate execution. The memory-bound regime corresponds to larger $N$, where global memory reads, writes, and normalization traffic dominate both time and energy consumption. 
Each term in the estimator captures a distinct component of causal softmax execution: $N$ represents the number of independent softmax normalizations (one per row); $N^2$ captures the dominant memory traffic and total computations of the attention matrix; $\log_2(N)$ models the reduction depth for normalization and associated synchronization overhead; and the bias accounts for fixed overheads such as kernel launch latency, instruction dispatch, and masking control flow, which dominate at small $N$.

Softmax is predominantly a memory-bound operator, with energy consumption driven primarily by quadratic memory access rather than floating-point computation. Separating kernel-launch dominated and memory-bound regimes prevents systematic error at small $N$ and captures bandwidth-driven scaling at larger $N$. 

\subsection{GEMM Estimator (Analytical)}
\label{subsec:gemm}
General matrix multiplication (GEMM) is a dominant building block of LLM inference, appearing in linear projections and feed-forward layers. On NVIDIA GPUs, these GEMMs are typically executed using \textbf{cuBLAS} (CUDA Basic Linear Algebra Subprograms), a closed-source, highly optimized library that selects among multiple internal algorithms depending on matrix dimensions, data type, and hardware constraints. Its execution time and energy consumption depend on several interacting factors, including the total number of operations, memory access patterns, arithmetic intensity, and the specific kernel implementation chosen by cuBLAS, such as different tiling strategies, tensor-core utilization, and multiplication schemes.

The cost of GEMM computation for \(C^{N \times M} = A^{N \times K} \times B^{K \times M}\) scales as $\mathcal{O}(MNK)$. We define the total floating-point workload as $F = 2MNK$.
The next factor is memory traffic, represented in bytes as $B_{\text{in}} = (M+N)K \cdot b \quad \& \quad B_{\text{out}} = MN \cdot b$.
\noindent where $b$ denotes bytes per element. 
We also consider the arithmetic intensity (AI) defined as $\text{AI} = \frac{F}{B_{\text{in}} + B_{\text{out}}}$, which quantifies the ratio of computation to memory traffic.

GEMM performance and energy exhibit distinct regimes depending on problem size and arithmetic intensity. We therefore partition GEMMs into five execution behaviors defined by thresholds on total work $F$ and arithmetic intensity (AI). The kernel-launch-bound regime corresponds to very small problems where $F < F_{\text{launch}}$. For all other regimes ($F \ge F_{\text{launch}}$), the behavior is determined by arithmetic intensity:

\begin{itemize}
    \item \textbf{kernel-launch-bound:} $F < F_{\text{launch}}$
    \item \textbf{memory-bound:} $\text{AI} < AI_{\text{low}}$
    \item \textbf{balanced (memory-dominant):} $AI_{\text{low}} \le \text{AI} < AI_{\text{mid}}$
    \item \textbf{balanced (compute-dominant):} $AI_{\text{mid}} \le \text{AI} < AI_{\text{high}}$
    \item \textbf{compute-bound:} $\text{AI} \ge AI_{\text{high}}$
\end{itemize}

This separation isolates kernel-launch overhead for small problems and distinguishes memory- and compute-dominated behavior at larger scales. The threshold parameters $F_{\text{launch}}$, $AI_{\text{low}}$, $AI_{\text{mid}}$, and $AI_{\text{high}}$ are defined as part of the category definitions above.

Beyond workload, memory traffic, and AI, GEMM time and energy depend strongly on achievable GPU utilization. In matrix multiplication, independent output elements can be computed in parallel in the $M$ and $N$ dimensions, whereas accumulation along the $K$ dimension is inherently serial due to the operation's structure. To capture this behavior, we define a parallelism factor as $PF = \frac{MN}{K}$. This reflects parallel work over $M,N$ relative to serialization over $K$. Larger PF indicates higher parallelism and utilization, while large $K$ increases serialization and reduces efficiency. The inverse term $\text{PF}^{-1}$ captures regimes where limited parallelism leads to higher execution cost.

The selected features were obtained via ablation, retaining only those with significant predictive contribution. For each regime, GEMM time and energy are modeled as:

\begin{equation}
\begin{aligned}
E_{\text{GEMM}} =
\gamma_{10}^{E} \text{PF}\cdot F 
+ \gamma_9^{E} \text{PF}\cdot B_{\text{in}} 
+ \gamma_8^{E} \text{AI}\cdot B_{\text{in}} 
+ \gamma_7^{E} F 
+ \gamma_6^{E} \text{PF} \\
+ \gamma_5^{E} \text{AI} 
+ \gamma_4^{E} B_{\text{out}}
+ \gamma_3^{E} B_{\text{in}}
+ \gamma_2^{E} \log_{10}(F)
+ \gamma_1^{E} \text{PF}^{-1}
+ \gamma_0^{E}
\end{aligned}
\end{equation}

\begin{equation}
\begin{aligned}
T_{\text{GEMM}} =
\gamma_{10}^{T} \text{PF}\cdot F 
+ \gamma_9^{T} \text{PF}\cdot B_{\text{in}} 
+ \gamma_8^{T} \text{AI}\cdot B_{\text{in}} 
+ \gamma_7^{T} F 
+ \gamma_6^{T} \text{PF} \\
+ \gamma_5^{T} \text{AI} 
+ \gamma_4^{T} B_{\text{out}}
+ \gamma_3^{T} B_{\text{in}}
+ \gamma_2^{T} \log_{10}(F)
+ \gamma_1^{T} \text{PF}^{-1}
+ \gamma_0^{T}
\end{aligned}
\end{equation}

Each term in the estimator captures a distinct component of GEMM execution:

\begin{itemize}

\item \textbf{$F$ and $\log_{10}(F)$:} Capture total compute and small-kernel effects such as launch overhead and scheduling inefficiencies.

\item \textbf{$B_{\text{in}}$ and $B_{\text{out}}$:} Represent dominant memory traffic from reading input and writing output.

\item \textbf{$\text{AI}$:} Captures the balance between computation and memory, distinguishing execution regimes.

\item \textbf{$\text{PF}$:} Represents the available parallelism in $M$ and $N$ relative to the depth of reduction in $K$, influencing GPU utilization.

\item \textbf{$\text{PF}^{-1}$:} Captures serialization effects for limited parallelism.

\item \textbf{Interaction terms ($\text{PF}\cdot F$, $\text{PF}\cdot B_{\text{in}}$, $\text{AI}\cdot B_{\text{in}}$):} Capture interactions between compute workload, memory traffic, and available parallelism. For example, $(\text{PF}\cdot F)$ reflects how additional computation scales differently depending on available parallelism.

\end{itemize}

\vspace{-3mm}

\subsection{Attention Block Estimator (ML-based)}
\label{subsec:attn}
The attention block is dominated by matrix multiplications and softmax operations. We therefore estimate its execution time and energy using the GEMM and Softmax base estimators. The GEMM estimator is applied to the linear projections that generate query ($Q$), key ($K$), and value ($V$), as well as to $QK^{T}$, the attention-weighted value $Softmax(QK^{T}) \cdot V$, and the output projection. The Softmax estimator normalizes the attention score matrix.

However, several operations are not explicitly captured, including tensor reshaping (e.g., head splitting and concatenation), masking, and other lightweight memory operations. While individually inexpensive, these collectively introduce non-negligible overhead.

To capture these residual costs, we employ a multilayer perceptron (MLP) correction model. It takes as input the predicted time and energy from GEMM and softmax estimators, along with architectural parameters such as number of heads, head dimension, and sequence length, and predicts total attention block time and energy. This enables modeling of both dominant kernel costs and attention-specific overheads.

The model supports both multi-head attention (MHA) and grouped-query attention (GQA). During autoregressive decoding, each forward pass processes a single newly generated token, so the operator-level query length is fixed at $N=1$; however, this does not remove the dependence on context length. Let $L_{\text{KV}}$ denote the KV-cache length. In the $QK^{T}$ multiplication, the GEMM dimensions are $(1,d_{\text{head}}) \times (d_{\text{head}},L_{\text{KV}})$, so the cache length appears as the output dimension. Similarly, for $Softmax(QK^{T})V$, the GEMM dimensions are $(1,L_{\text{KV}}) \times (L_{\text{KV}},d_{\text{head}})$, where $L_{\text{KV}}$ appears as the reduction dimension. Thus, increasing context length is explicitly modeled through the GEMM dimensions and the corresponding softmax input size, even though only one new token is processed per decoding step. During prefill, the estimators use the full prompt length. The attention estimator is implemented as an MLP with $L_{\text{attn}}$ layers and hidden dimension $D_{\text{attn}}$.
\vspace{-2mm}
\subsection{FFN Block Estimator (ML-based)}
\label{subsec:ffn}

The feed-forward network (FFN) block is dominated by dense matrix multiplications. In gated variants, it consists of two parallel up projections followed by a down projection. We estimate their execution time and energy using the GEMM estimator, applied to the up projections (expanding to the intermediate dimension) and the down projection (mapping back to the model dimension).

Between these projections, the FFN includes lightweight operations such as activations (e.g., Gaussian error linear unit (GELU) and sigmoid linear unit (SiLU)), element-wise multiplications, and intermediate memory accesses. Although small relative to GEMMs, they introduce additional overhead not captured by the base estimators.

To account for this, we use the same MLP-based correction model as in the attention block. It takes as input the predicted time and energy of the projections, along with architectural parameters such as model dimension ($d_{\text{model}}$) and intermediate dimension ($d_{\text{intermediate}}$), and predicts total FFN execution time and energy.

The formulation also supports MoE architectures. Since experts share the same structure, base estimators are applied to a representative expert, while the number of experts, top-$k$ routing, and shared expert dimension (if present) are provided as additional MLP inputs to predict total MoE FFN cost.

During decoding, estimators use $N=1$, while during prefill, they use the prompt length. The FFN estimator is implemented as an MLP with $L_{\text{ffn}}$ layers and hidden dimension $D_{\text{ffn}}$.
\vspace{-2mm}
\subsection{End-to-End Estimator (ML-based)}
\label{subsec:fixator}

Using the operator- and block-level models, we first estimate the execution time and energy of the dominant components in each LLM layer: RMSNorm, attention, and FFN, which account for most computation and memory traffic during inference. These predictions are then fed into a final ML-based estimator, together with global features such as model dimension ($d_{\text{model}}$), number of layers, sequence length, batch size, inference mode (decode vs.\ prefill), and MoE configuration.

This estimator predicts total LLM inference time and energy, capturing system-level effects not modeled at lower levels. In particular, this stage is harder to predict due to complex interactions between layers, including lightweight operations between components and runtime overheads such as kernel launches, synchronization, and cross-layer scheduling. By combining block-level predictions with global features, it provides accurate end-to-end estimates. The estimator is implemented as a lightweight MLP with $L_{\text{final}}$ layers and hidden dimension $D_{\text{final}}$.

\vspace{-2mm}
\section{Applications in Optimization}
The proposed framework enables efficient exploration of the LLM architectural design space by directly predicting execution time and energy from architectural parameters, serving as a fast surrogate for hardware evaluation.

One key application is architecture-level optimization for energy or latency. Given a model configuration, the estimator evaluates variations in parameters such as the number of attention heads, key-value heads, model dimension ($d_{\text{model}}$), feed-forward dimension ($d_{\text{intermediate}}$), number of layers, and MoE settings. By sweeping these parameters, it identifies configurations that minimize energy or runtime.

\begin{figure}[t]
\centering
\begin{subfigure}{0.49\columnwidth}
\centering
\includegraphics[width=\linewidth]{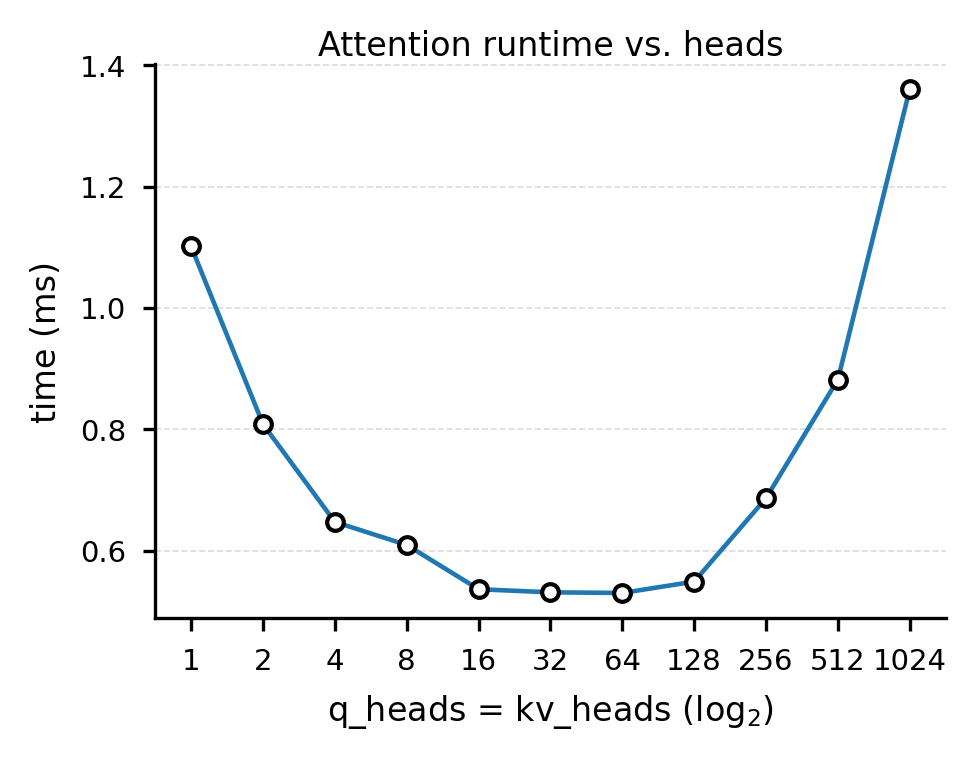}
\caption{Attention runtime vs. heads}
\end{subfigure}
\hfill
\begin{subfigure}{0.49\columnwidth}
\centering
\includegraphics[width=\linewidth]{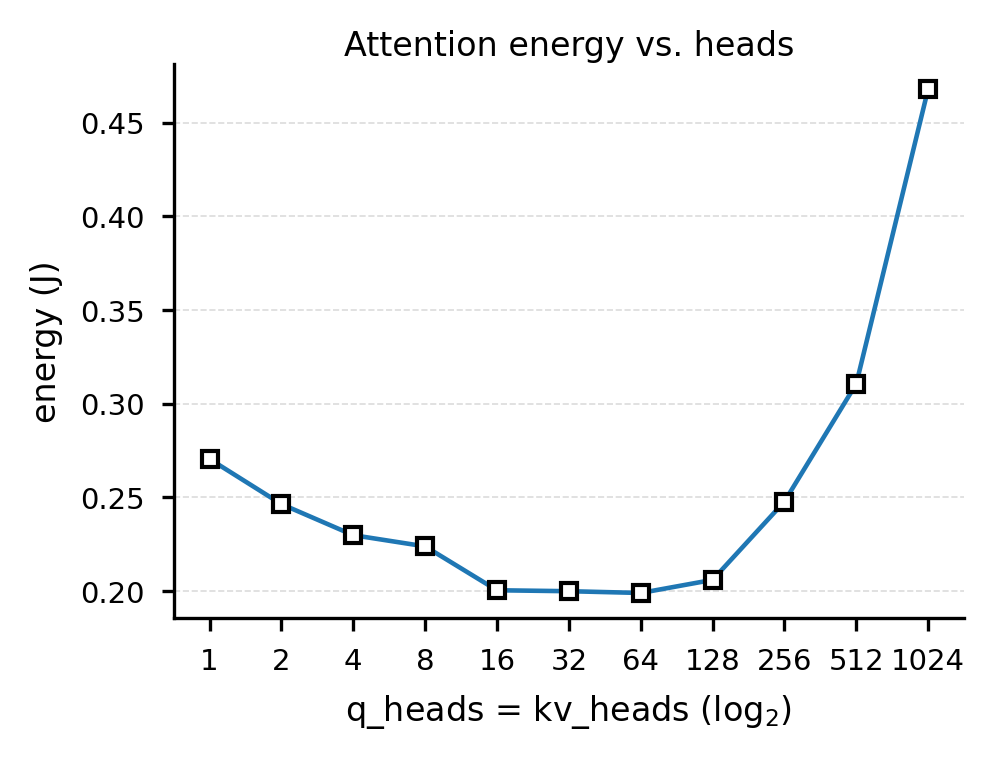}
\caption{Attention energy vs. heads}
\end{subfigure}

\caption{Effect of the number of attention heads on execution time and energy for $d_{\text{model}}=4096$ and $q_{\text{heads}}=kv_{\text{heads}}$.}
\label{fig:head_opt}
\end{figure}

As an example, we vary the number of attention heads while fixing $d_{\text{model}}=4096$ and using equal query and key-value heads. For a single attention layer in prefill ($N=1024$), Fig.~\ref{fig:head_opt} shows that $64$ heads achieve the lowest energy, followed by $32$, while $16$ and $128$ remain competitive. Very large headcounts increase time and energy due to reduced parallel efficiency and higher overhead.

More broadly, the estimator supports joint optimization across parameters. For instance, increasing $d_{\text{model}}$ while reducing layers can preserve capacity while improving efficiency. Similarly, jointly tuning $d_{\text{model}}$, $d_{\text{intermediate}}$, attention heads, and key-value heads enables hardware-aware designs. This approach applies at both block level (attention/FFN) and full-model level for energy- or latency-optimized LLM architectures.

\section{EXPERIMENTS AND RESULTS}

\subsection{Data Collection and Measurements}
Time was measured using CUDA events, and energy was measured with NVML by reading the GPU energy counter before and after each measured region. To reduce noise, each operator or configuration was repeated until the cumulative execution time exceeded 10 seconds; per-execution latency and energy were then obtained by dividing the total time and energy by the number of repetitions. For large prefill workloads whose individual executions were already long, we used five repetitions, since the longer measurement window makes NVML granularity noise negligible. All experiments were conducted in bfloat16 (BF16) precision. To avoid cache reuse effects and reflect realistic memory behavior, we rotated pre-generated weight sets across runs. We constructed datasets that span diverse architectural configurations (varying model dimensions, layers, sequence lengths, batch sizes, dense/MoE FFNs, and MHA/GQA) for both prefill and decode modes. The end-to-end evaluation used a dataset of approximately 1,200 model configurations profiled with HuggingFace Transformers.

We trained MLP-based regressors for attention, FFN, and full-model latency and energy prediction. Inputs and targets were standardized using training-set statistics. All estimators used 4-hidden-layer MLPs. The attention estimators used 384 hidden units and Smooth L1 loss, optimized with AdamW using a learning rate of $5\times10^{-5}$ and weight decay of $10^{-5}$ for 2,000 epochs. The FFN estimators used width 64 and MSE loss, optimized with AdamW using a learning rate of $2\times10^{-3}$ and weight decay of $10^{-6}$ with early stopping. The full-model prefill and decode estimators used width 32 and MSE loss on log-transformed latency and energy, optimized with AdamW using a learning rate of $2\times10^{-3}$ and weight decay of $10^{-6}$ for 100 epochs. We trained the full-model estimators in log-space because this objective better matches relative-error evaluation and produced the lowest MAPE in our experiments.

While the current evaluation is conducted on an NVIDIA H100 NVL GPU, HYMELL's hierarchical design is fundamentally hardware-portable. Adapting the framework to new architectures (e.g., earlier GPU generations or alternative accelerators) does not require redesigning the architecture; it only necessitates re-profiling the hardware to update the analytical regression coefficients (e.g., $\alpha, \beta, \gamma$) and fine-tuning the lightweight MLPs on the new target's data.
Because ML estimators are implemented as lightweight MLPs ($L_{\text{attn}}$, $L_{\text{ffn}}$, $L_{\text{final}}$) operating on low-dimensional architectural features rather than raw tensor data, their training overhead is trivial. Furthermore, generating the 1,200 profiling configurations is a one-time automated hardware cost that requires significantly less manual engineering effort than re-deriving structural equations for purely analytical models \cite{cavagna2026sweetspotanalyticalmodelpredicting, patwari2025forecastingllminferenceperformance}.

We report prediction quality using mean absolute percentage error (MAPE) and the coefficient of determination ($R^2$).

\begin{table}[t]
\centering
\caption{Accuracy of the RMSNorm estimator.}
\vspace{-4mm}
\label{tab:rmsnorm_results}
\begin{tabular}{lcccc}
\toprule
 & \multicolumn{2}{c}{Time} & \multicolumn{2}{c}{Energy} \\
\cmidrule(r){2-3} \cmidrule(r){4-5}
Regime & MAPE (\%) & $R^2$ & MAPE (\%) & $R^2$ \\
\midrule
Launch-bound & 1.288 & 0.240156 & 3.233 & 0.979766 \\
Memory-bound & 2.098 & 0.999759 & 3.516 & 0.999451 \\
\midrule
Average & 1.693 & 0.619958 & 3.375 & 0.989609 \\
\bottomrule
\end{tabular}
\end{table}

\begin{table}[t]
\centering
\caption{Accuracy of the Softmax estimator.}
\vspace{-4mm}
\label{tab:softmax_results}
\begin{tabular}{lcccc}
\toprule
 & \multicolumn{2}{c}{Time} & \multicolumn{2}{c}{Energy} \\
\cmidrule(r){2-3} \cmidrule(r){4-5}
Regime & MAPE (\%) & $R^2$ & MAPE (\%) & $R^2$ \\
\midrule
Launch-bound & 2.869 & 0.996185 & 2.583 & 0.999042 \\
Memory-bound & 3.661 & 0.999982 & 1.251 & 0.999970 \\
\midrule
Average & 3.265 & 0.998084 & 1.917 & 0.999506 \\
\bottomrule
\end{tabular}
\end{table}
\vspace{-2mm}
\subsection{Evaluation}
Tables~\ref{tab:rmsnorm_results} and ~\ref{tab:softmax_results}  report the accuracy of operator-level estimators across execution regimes. The estimators achieve high accuracy, with MAPE below $4\%$ for Softmax and below $4\%$ for RMSNorm, demonstrating the effectiveness of HYMELL.

In the launch-bound regime of RMSNorm, execution time is dominated by kernel launch overhead and remains nearly constant across configurations. This results in a low variance in ground-truth measurements and therefore a lower $R^2$, despite a low prediction error. The regime thresholds are $th_{\text{Softmax}}=1024$ for softmax and $th_{\text{RMS}} = 2.5 \times 10^{6}$ for RMSNorm.

\begin{table}[t]
\centering
\caption{Accuracy of the GEMM estimator.}
\vspace{-4mm}
\label{tab:gemm_results}
\begin{tabular}{lccccc}
\toprule
 & \multicolumn{2}{c}{Time} & \multicolumn{2}{c}{Energy} & \\
\cmidrule(r){2-3} \cmidrule(r){4-5}
Regime & MAPE (\%) & $R^2$ & MAPE (\%) & $R^2$ \\
\midrule
Launch-bound        & 5.587 & 0.834 & 6.134 & 0.930 \\
Memory-bound        & 6.819 & 0.993 & 5.861 & 0.996  \\
Balanced (mem-dom)  & 5.974 & 0.993 & 5.189 & 0.992 \\
Balanced (comp-dom) & 9.399 & 0.960 & 7.294 & 0.979 \\
Compute-bound       & 4.141 & 0.999 & 4.472 & 0.998  \\
\midrule
Average             & 6.384 & 0.956 & 5.790 & 0.979 \\
\bottomrule
\end{tabular}
\end{table}

Table~\ref{tab:gemm_results} reports the accuracy of the GEMM estimator in execution regimes, with MAPE generally below $7\%$ for both time and energy, indicating stable performance in diverse workload characteristics.
Regimes are defined by total work $F$ and arithmetic intensity (AI): $F_{\text{thr}} = 8.39\times10^{6}$ floating-point operations (FLOPs), $\text{AI}_{low} = 81.92$, $\text{AI}_{mid} = 166.05$, and $\text{AI}_{high} = 288.91$ FLOPs/byte.
Higher errors in balanced regimes stem from transitions between memory- and compute-bound behavior, where GPU utilization and scheduling vary. Additionally, cuBLAS dynamically selects kernels and tiling strategies based on input dimensions and AI, introducing performance discontinuities that make these regimes harder to model.

Tables~\ref{tab:attention_results} and~\ref{tab:ffn_results} report the accuracy of ML-based estimators for attention and FFN. Both achieve high accuracy, with MAPE below $3\%$ for attention and below $6\%$ for FFN.
For attention, prefill exhibits higher $R^2$ since computation scales with sequence length and dominant GEMM and Softmax operations are well modeled. During decoding, each inference step computes only the newly generated token while reusing cached keys and values from previous tokens. Consequently, the computational workload outside the attention kernel remains effectively constant across decoding steps. Although attention still depends on the KV-cache length, on modern GPUs this dependence is largely memory-bandwidth dominated and grows much more slowly than the prefill workload. As a result, the measured decode latency exhibits substantially lower variance than prefill across the evaluated context lengths, which explains the lower $R^2$ despite the very low MAPE.
For FFN, the estimator accurately captures both dense and MoE architectures. Slightly higher error in MoE arises from additional routing and memory overhead for expert selection and aggregation.
Overall, these results show that the ML-based block estimators effectively capture residual computation and memory overhead beyond the analytical operator models.
\begin{table}[t]
\centering
\caption{Accuracy of the attention block estimator.}
\vspace{-4mm}
\label{tab:attention_results}
\begin{tabular}{lcccc}
\toprule
 & \multicolumn{2}{c}{Time} & \multicolumn{2}{c}{Energy} \\
\cmidrule(r){2-3} \cmidrule(r){4-5}
Mode & MAPE (\%) & $R^2$ & MAPE (\%) & $R^2$ \\
\midrule
Prefill & 3.514 & 0.9997 & 4.237 & 0.9996 \\
Decode  & 0.441 & 0.9276 & 0.599 & 0.9996 \\
\midrule
Average & 1.978 & 0.9636 & 2.418 & 0.9996 \\
\bottomrule
\end{tabular}
\end{table}

\begin{table}[t]
\centering
\caption{Accuracy of the FFN block estimator.}
\vspace{-4mm}
\label{tab:ffn_results}
\begin{tabular}{lcccc}
\toprule
 & \multicolumn{2}{c}{Time} & \multicolumn{2}{c}{Energy} \\
\cmidrule(r){2-3} \cmidrule(r){4-5}
Architecture & MAPE (\%) & $R^2$ & MAPE (\%) & $R^2$ \\
\midrule
Dense FFN & 4.063 & 0.9998 & 4.612 & 0.9998 \\
MoE FFN   & 5.168 & 0.9946 & 6.333 & 0.9951 \\
\midrule
Average   & 4.616 & 0.9972 & 5.473 & 0.9975 \\
\bottomrule
\end{tabular}
\end{table}

Table~\ref{tab:e2e_results} reports end-to-end prediction accuracy on held-out LLM architectures. The model achieves $\sim$10\% MAPE for both time and energy across dense and MoE models. Higher error compared to operator- and block-level estimators is expected, as end-to-end prediction must capture additional effects such as kernel scheduling, cross-layer memory interactions, and framework overheads. Nevertheless, high $R^2$ values indicate accurate modeling of performance trends across diverse architectures. 
Furthermore, while memory optimizations like PageAttention \cite{kwon2023efficientmemorymanagementlarge} and FlashAttention \cite{dao2022flashattentionfastmemoryefficientexact, dao2023flashattention2fasterattentionbetter,
shah2024flashattention3fastaccurateattention} alter memory layout and improve utilization to enable larger batch sizes, they do not fundamentally change the computational workload. Because batch size is explicitly modeled as an input feature in HYMELL, our framework naturally captures the performance and energy implications of these optimizations without requiring changes to the underlying estimators.

While our evaluation focuses on uniform batch configurations, modern serving engines employ continuous batching with heterogeneous sequence lengths. In such settings, multiple requests are processed jointly within shared kernels, leading to non-trivial interactions in computation and memory access patterns. Since HYMELL models operator costs as a function of input dimensions, it can be extended to heterogeneous batches by evaluating the effective aggregated workload at the kernel level rather than simply summing independent requests. This allows the model to approximate the combined execution cost while accounting for shared computation and improved hardware utilization.

\begin{table}[t]
\centering
\caption{End-to-end LLM prediction accuracy.}
\vspace{-4mm}
\label{tab:e2e_results}
\begin{tabular}{lcccc}
\toprule
 & \multicolumn{2}{c}{Time} & \multicolumn{2}{c}{Energy} \\
\cmidrule(r){2-3} \cmidrule(r){4-5}
Configuration & MAPE (\%) & $R^2$ & MAPE (\%) & $R^2$ \\
\midrule
Prefill (Dense) & 10.179 & 0.9938 & 9.969 & 0.9860 \\
Decode (Dense)  & 8.194  & 0.8968 & 10.771 & 0.8117 \\
Prefill (MoE)   & 12.772 & 0.9527 & 13.091 & 0.9477 \\
Decode (MoE)    & 12.776 & 0.9221 & 13.305 & 0.9101 \\
\midrule
Average         & 10.980 & 0.9414 & 11.784 & 0.9139 \\
\bottomrule
\end{tabular}
\end{table}

\begin{table}[t]
\centering

\caption{HYMELL per-sequence accuracy compared with AMALI’s reported aggregate MAPE on LLaMA 3 8B with batch size 1.}
\vspace{-4mm}
\label{tab:amali_compare_full}
\resizebox{\linewidth}{!}{%
\begin{tabular}{c|c|cc|cc}
\toprule
\textbf{Mode} & \textbf{Seq Len} & \multicolumn{2}{c|}{\textbf{Time Error (\%)}} & \multicolumn{2}{c}{\textbf{Energy Error (\%)}} \\
 &  & \textbf{AMALI} & \textbf{HYMELL} & \textbf{AMALI} & \textbf{HYMELL} \\
\midrule
\multirow{5}{*}{\parbox[c]{1.2cm}{\centering Prefill}}
& 256  & 15.56 & 5.65 & -- & 2.60 \\
& 1024 & 15.56 & 4.44 & -- & 4.03 \\
& 2048 & 15.56 & 1.60 & -- & 1.67 \\
& 4096 & 15.56 & 5.09 & -- & 3.38 \\
& \textbf{MAPE} & 15.56 & 4.19 & -- & 2.92 \\
\midrule
\multirow{5}{*}{\parbox[c]{1.2cm}{\centering Decode}}
& 256  & 34.90 & 1.64 & -- & 2.04 \\
& 1024 & 34.90 & 1.16 & -- & 1.22 \\
& 2048 & 34.90 & 1.79 & -- & 1.09 \\
& 4096 & 34.90 & 1.71 & -- & 1.00 \\
& \textbf{MAPE} & 34.90 & 1.58 & -- & 1.34 \\
\bottomrule
\end{tabular}%
}
\end{table}

Table~\ref{tab:amali_compare_full} reports HYMELL’s per-sequence prediction error for LLaMA 3 8B with batch size 1 and compares it with AMALI’s reported aggregate MAPE for prefill and decode. AMALI reports cycle-prediction MAPE of $15.56\%$ for prefill and $34.90\%$ for decode, but does not provide per-sequence errors or energy estimates. Therefore, the AMALI values are repeated across sequence lengths only as an aggregate reference baseline, not as per-sequence measurements. In contrast, HYMELL predicts measured wall-clock latency and energy for each sequence length, achieving $4.19\%$ time MAPE and $2.92\%$ energy MAPE for prefill, and $1.58\%$ time MAPE and $1.34\%$ energy MAPE for decode. Unlike AMALI’s analytical cycle-level model, HYMELL predicts measured hardware latency and energy, capturing runtime overheads, scheduling effects, and frequency behavior while also enabling system-level energy prediction.

\begin{figure}[t]
\centering
\includegraphics[width=\linewidth]{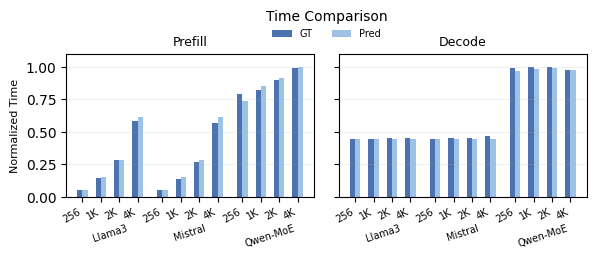}
\vspace{-3mm}
\includegraphics[width=\linewidth]{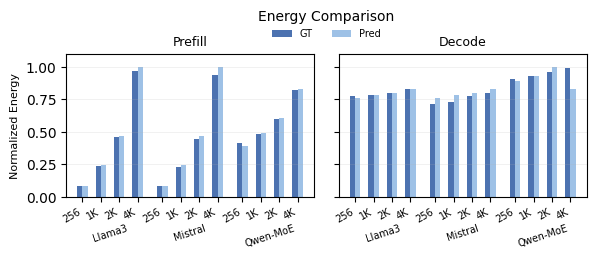}
\caption{Normalized prediction accuracy (batch=1). Top: time (ground truth normalized to 1). Bottom: energy (ground truth normalized to 1).}
\label{fig:accuracy_combined}
\end{figure}

Figure~\ref{fig:accuracy_combined} presents the normalized prediction accuracy of HYMELL for execution time and energy across three representative models—LLaMA~3~8B \cite{grattafiori2024llama}, Mistral~7B \cite{jiang2023mistral7b}, and Qwen~1.5~2.7B MoE \cite{qwen} (2.7B active parameters)—over different sequence lengths with batch size 1. Results are shown for both prefill and decode stages, where ground-truth values are normalized to one, and predictions are plotted relative to them. Across all configurations, predictions closely match the ground truth, with errors below $5\%$, demonstrating that HYMELL accurately captures both compute- and memory-dominated behaviors. The consistent accuracy across diverse architectures and workloads highlights the robustness and generalizability of the proposed model.

\subsection{Generalization and Extensibility}
To further validate that HYMELL is not tied to a single hardware platform, we repeated the model-level evaluation on an NVIDIA RTX A6000 GPU. This experiment serves as a cross-device ablation: the same hierarchical modeling flow is retained, while the hardware-specific profiling data and learned coefficients are updated for the A6000. As shown in Table~\ref{tab:a6000_ablation}, HYMELL achieves strong accuracy on the new device for both dense and MoE models, with average MAPE values of $8.141\%$ for time and $8.622\%$ for energy. These results demonstrate that the proposed analytical--ML decomposition transfers well across GPU generations once the target hardware is profiled, supporting the hardware-portable design of the framework.

\begin{table}[t]
\centering
\caption{Cross-device ablation on an NVIDIA RTX A6000 GPU.}
\vspace{-4mm}
\label{tab:a6000_ablation}
\begin{tabular}{lcccc}
\toprule
 & \multicolumn{2}{c}{Time} & \multicolumn{2}{c}{Energy} \\
\cmidrule(r){2-3} \cmidrule(r){4-5}
Configuration & MAPE (\%) & $R^2$ & MAPE (\%) & $R^2$ \\
\midrule
Prefill (Dense) & 9.102 & 0.9814 & 8.425 & 0.9805 \\
Decode (Dense)  & 6.718 & 0.9742 & 7.342 & 0.9747 \\
Prefill (MoE)   & 9.366 & 0.9734 & 9.367 & 0.9728 \\
Decode (MoE)    & 7.379 & 0.9813 & 9.352 & 0.9379 \\
\midrule
Average         & 8.141 & 0.9776 & 8.622 & 0.9665 \\
\bottomrule
\end{tabular}
\end{table}

We also evaluate HYMELL on a linear-attention variant to test architectural extensibility beyond the softmax-based attention blocks used in most current LLMs. Although linear attention is not the dominant attention mechanism in contemporary LLM deployments, it is an important example of a structurally different attention formulation. Table~\ref{tab:linear_attention_ablation} shows that HYMELL can estimate both latency and energy for this variant with low error, achieving average MAPE values of $1.910\%$ for time and $2.278\%$ for energy. This confirms that new attention mechanisms can be incorporated by adding the corresponding block-level profiling data and estimator, without changing the overall multi-level modeling framework.

\begin{table}[t]
\centering
\caption{Architectural ablation on linear attention.}
\vspace{-4mm}
\label{tab:linear_attention_ablation}
\begin{tabular}{lcccc}
\toprule
 & \multicolumn{2}{c}{Time} & \multicolumn{2}{c}{Energy} \\
\cmidrule(r){2-3} \cmidrule(r){4-5}
Configuration & MAPE (\%) & $R^2$ & MAPE (\%) & $R^2$ \\
\midrule
Prefill & 3.360 & 0.9980 & 3.740 & 0.9958 \\
Decode  & 0.460 & 0.9999 & 0.815 & 0.9999 \\
\midrule
Average & 1.910 & 0.9990 & 2.278 & 0.9979 \\
\bottomrule
\end{tabular}
\end{table}

Currently, HYMELL focuses on single-GPU inference execution. However, the framework's block-level modularity naturally lends itself to multi-GPU scaling. To model large-scale models that require tensor parallelism (TP), the framework can be extended by introducing an additional analytical estimator for collective communication primitives and by incorporating network bandwidth as a global feature in the end-to-end MLP predictor.
\vspace{-2mm}
\section{CONCLUSIONS}
We introduced HYMELL, a hybrid multi-level framework for predicting LLM inference latency and energy consumption. By combining regime-aware analytical modeling for GPU operators (GEMM, Softmax, RMSNorm) with machine learning predictors for higher-level components and full-system execution, HYMELL bridges low-level hardware behavior and high-level architecture. Evaluated on NVIDIA H100 GPUs, HYMELL achieves high predictive accuracy (MAPE of 1.9\%--6.3\% at block levels and less than 5\% for Llama3, Mistral, and Qwen models) across diverse configurations, including GQA and MoE paradigms in both prefill and decode phases. By predicting execution costs directly from architectural parameters, HYMELL enables rapid, hardware-free design space exploration, serving as an adaptable foundation for energy-efficient AI deployment.

\end{document}